\pdfoutput=1

\documentclass[11pt]{article}

\usepackage{authblk}
\usepackage[]{EACL2023}

\makeatletter
\renewcommand\AB@affilsepx{, \protect\Affilfont}
\makeatother

\usepackage{amsmath}
\usepackage{times}
\usepackage{graphicx}
\usepackage{xcolor}
\usepackage{pdfpages}
\usepackage{float}
\usepackage{booktabs}
\usepackage{tabulary}
\usepackage{latexsym}
\usepackage{hyperref}
\usepackage{multirow}
\usepackage{multicol}
\usepackage{verbatim}

\usepackage[T1]{fontenc}

\usepackage[utf8]{inputenc}

\usepackage{microtype}

\usepackage{inconsolata}

\title{\textcolor{red}{Red}HOT: A Corpus of Annotated Medical Questions, Experiences, and Claims on Social Media}

\author{\textbf{Somin Wadhwa}$^{\dagger}$\quad \textbf{Vivek Khetan}$^{\Diamond}$\quad \textbf{Silvio Amir}$^{\dagger}$\quad \textbf{Byron C. Wallace}$^{\dagger}$\quad \\ 
Northeastern University$^{\dagger}$ \quad Accenture AI  Labs$^{\Diamond}$ \\ 
\texttt{\{wadhwa.s,s.amir,b.wallace\}@northeastern.edu} \\ 
\texttt{vivek.a.khetan@accenture.com}
}

\begin{document}
\maketitle


\begin{abstract}
We present \textbf{\textcolor{red}{Red}}dit \textbf{H}ealth \textbf{O}nline \textbf{T}alk (\textbf{\textcolor{red}{Red}HOT}), a corpus of 22,000 richly annotated social media posts from Reddit spanning 24 health conditions. 
Annotations include demarcations of spans corresponding to medical claims, personal experiences, and questions.
We collect additional granular annotations on identified claims.
Specifically, we mark snippets that describe patient \textbf{P}opulations, \textbf{I}nterventions, and \textbf{O}utcomes (PIO elements) within these. 
Using this corpus, we introduce the task of retrieving trustworthy evidence relevant to a given claim made on social media. 
We propose a new method to automatically derive (noisy) supervision for this task which we use to train a dense retrieval model; this outperforms baseline models. 
Manual evaluation 
of retrieval results performed by medical doctors indicate that while our system performance is promising, there is considerable room for improvement.
We 
release all annotations collected (and scripts to assemble the dataset), and all code necessary to reproduce the results in this paper at: \url{https://sominw.com/redhot}.  
\end{abstract}

\section{Introduction}
Social media platforms 
such as 
Reddit provide individuals places to discuss (potentially rare) medical conditions that affect them. 
This allows people to communicate with others who share in their condition, exchanging 
information about symptom trajectories, personal experiences, and treatment options. 
Such communities can provide support \cite{biyani2014identifying} and access to information about rare conditions which may otherwise be difficult to find \cite{glenn2015using}. 

However, the largely unvetted nature of social media platforms make them vulnerable to \textit{mis} and \textit{disinformation} \cite{swire2019public}.
An illustrative and timely example is the idea that consuming bleach might be a viable treatment for COVID-19,\footnote{\url{https://www.theguardian.com/world/2020/sep/19/bleach-miracle-cure-amazon-covid}} which quickly gained traction on social media. 
All misinformation can be dangerous, but \textit{medical} misinformation poses unique risks to public health, especially as individuals increasingly turn to social media to inform personal health decisions \cite{nobles2018std,BARUA2020100119}.

\begin{figure}
    \centering
\includegraphics[scale=0.36]{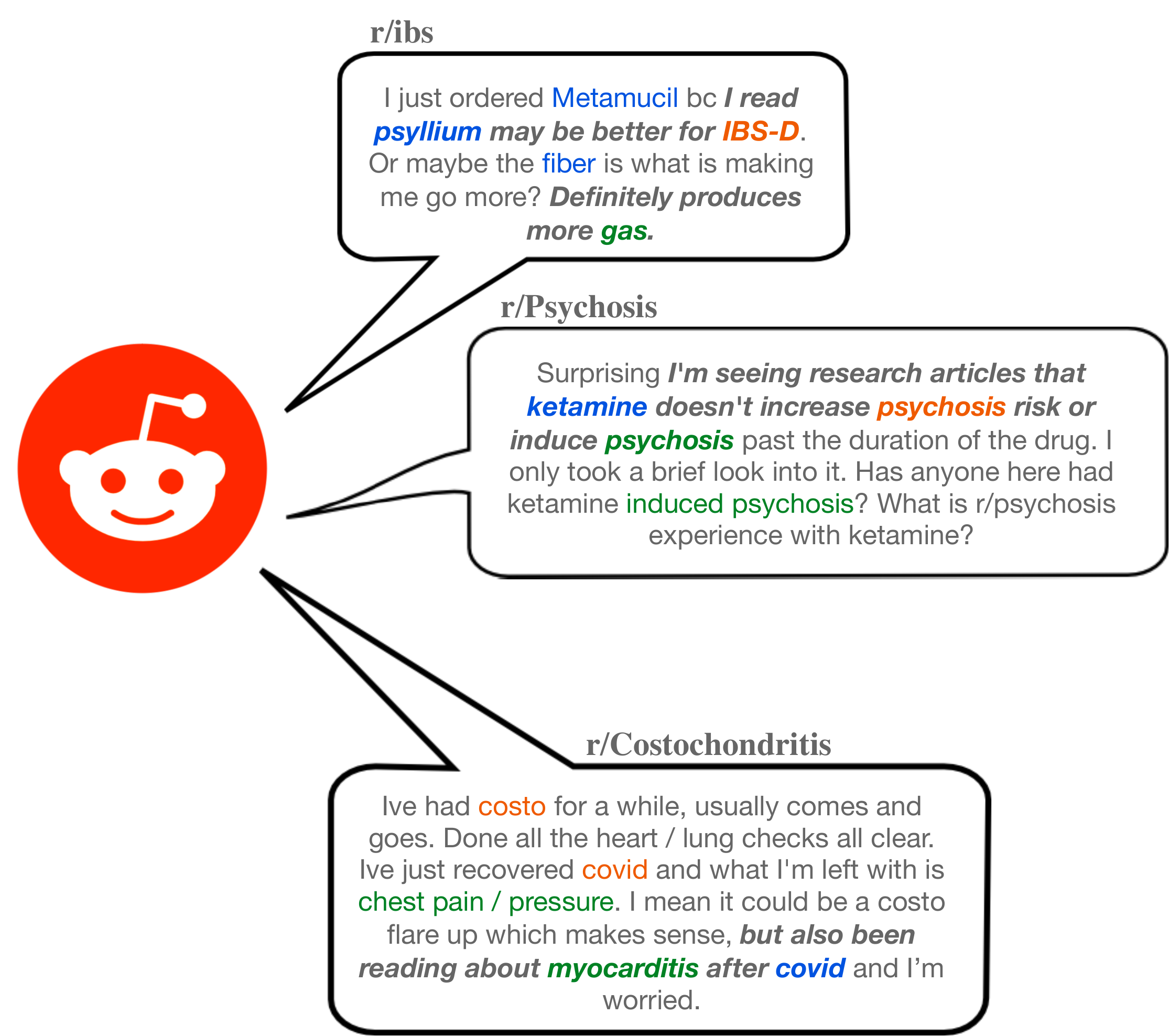}
\caption{Examples of health-related Reddit posts annotated for populations, interventions, and outcomes.}
\label{fig:overview}
\end{figure}

In this paper, we introduce \textbf{\textcolor{red}{Red}HOT}: an annotated dataset of health-related claims, questions, and personal experiences posted to Reddit. 
This dataset can support development of a wide range of models for processing health-related posts from social media.  
Unlike existing health-related social media corpora, \textbf{\textcolor{red}{Red}HOT}: (a) Covers a broad range of health topics (e.g., not just COVID-19), and, (b) Comprises ``natural'' claims collected from real health-related fora (along with annotated questions and personal experiences). 
Furthermore, we have collected granular annotations on claims, demarcating descriptions of the \textcolor{orange}{Population} (e.g., \textcolor{orange}{diabetics}), \textcolor{blue}{Interventions}, and \textcolor{teal}{Outcomes}, i.e., the \emph{PIO} elements \cite{richardson1995well}. 
Such annotations may permit useful downstream processing: For example, in this work we use them to facilitate retrieval of evidence relevant to a claim. 

Specifically, we develop and evaluate a pipeline to automatically identify and contextualize health-related claims on social media, as we anticipate that such a tool might be useful for moderators keen to keep their communities free of potentially harmful misinformation. 
With this use-case in mind, we propose methods for automatically retrieving \emph{trustworthy} published scientific evidence relevant to a given claim made on social media, which may in aggregate support or debunk a particular claim. 

The contributions of this work are summarized as follows. 
First, we introduce \textbf{\textcolor{red}{Red}HOT}: A new dataset comprising $22,000$ health-related Reddit posts across 24 medical conditions annotated for claims, questions, and personal experiences. 
Claims are additionally annotated with 
 PIO elements. 
Second, we introduce the task of identifying health-related claims on social media, extracting the associated PIO elements, and then retrieving relevant and trustworthy evidence to support or refute such claims. 
Third, we propose \textbf{\textcolor{red}{Red}HOT}-DER, a Dense Evidence Retriever trained with heuristically derived supervision to retrieve medical literature relevant to health-related claims made on social media.
We evaluate baseline models for the first two steps on the \textbf{\textcolor{red}{Red}HOT} dataset and assess the retrieval step with relevance judgments collected from domain experts (medical doctors).

The Reddit posts we have collected are public and typically made under anonymous pseudonyms, but nonetheless these are health-related comments and so inherently sensitive. To respect this, we (a) notified all users in the dataset of their (potential) inclusion in this corpus, and provided opportunity to opt-out, and, (b) we do not release the data directly, but rather a script to download annotated comments, so that individuals may choose to remove their comments in the future. Furthermore, we consulted with our Institutional Review Board (IRB) and confirmed that the initial collection and annotation of such data does not constitute human subjects research. However, EACL reviewers rightly pointed out that certain \emph{uses} of this data may be sensitive. Therefore, to access the collected dataset we require researchers to self-attest that they have obtained prior \textit{approval} from their own IRB regarding their intended use of the corpus. 




\begin{table*}
\footnotesize
\centering
\begin{tabular}{p{6cm}p{4.6cm}p{2.6cm}}
\hline
\multicolumn{1}{c}{\textbf{Reddit post}} & \multicolumn{1}{c}{\textbf{Span labels}} & \multicolumn{1}{c}{\textbf{PIO elements from claims}} \\ \hline
\textit{I've seen a bunch of posts on here from people who say that \textcolor{blue}{glycopyrrolate} suddenly isn't working anymore for \textcolor{orange}{hyperhidrosis}}. I'm one of those person who has been facing this for a while now. Just wondering if anyone fixed it? Can't really ask my GP about it since he didn't even  know the meds existed. He just prescribed them for me when I asked for it & {\tt{Claim:}} I've seen a bunch of posts on here from people who say that glycopyrrolate suddenly isn't working anymore for Hyperhidrosis \newline {\tt{Question}:} Just wondering if anyone fixed it? & \textbf{P} hyperhidrosis \newline \textbf{I} glycopyrrolate  \\ 
\hline
 \textit{so i recently read that \textcolor{blue}{adderall} can trigger a psychotic break} $\&$ i was prescribed adderall years ago for my \textcolor{orange}{adhd} but now i just have constant \textcolor{teal}{hallucination episodes}. anyone else experience adderall induced psychosis?  & {\tt{Claim}:} so i recently read that adderall can trigger a psychotic break \newline {\tt{Personal Experience}:} i was prescribed adderall years ago for my adhd but now i just have constant hallucination episodes \newline {\tt{Question}:} anyone else experience adderall induced psychosis? & \textbf{P} adhd \newline \textbf{I} adderall \newline \textbf{O} hallucinations \\ 
 \hline
I've had \textcolor{orange}{costochondritis} for a while, usually comes and goes. Done all the heart/lung checks all clear.
I've just recovered \textcolor{blue}{covid} and what I'm left with is \textcolor{teal}{chest pain}/pressure. I mean it could be a costo flare up which makes sense, but also \textit{been reading about \textcolor{teal}{myocarditis} after \textcolor{blue}{covid}} and I'm worried, how can I tell which is which? & {\tt{Claim}:} been reading about myocarditis after covid \newline {\tt{Personal Experience}:} I'm left with is chest pain/pressure \newline {\tt{Question}:} how can I tell which is which? & \textbf{P} costochondritis \newline \textbf{I} covid \newline \textbf{O} myocarditis, chest-pain\\
 \hline
\end{tabular}
\centering
\caption{Example annotations, which include: extracted spans (phase 1), and spans describing \textcolor{orange}{\textbf{P}opulations}, \textcolor{blue}{\textbf{I}nterventions}, and \textcolor{teal}{\textbf{O}utcomes} --- PIO elements --- within them (phase 2). 
We collect the latter only for claims.
}

\label{tab:examples}
\end{table*}

\section{The \textcolor{red}{Red}HOT Dataset}
\label{section:data}


We have collected and manually annotated health related posts from Reddit to support development of language technologies which might, e.g., flag potentially problematic claims for moderation. 
Reddit is a social media platform that 
allows users to create their own communities (\textit{subreddits}) focused on specific topics. 
Subreddits are often about niche topics, and this permits in-depth discussion catering to a long tail of interests and experiences.
Notably, subreddits exist for most common (and many rare) medical conditions; we can therefore sample posts from such communities for annotation. 

\subsection{Data Annotation}

We decomposed data annotation into two stages, performed in sequence. 
In the first, workers are asked to demarcate spans of text corresponding to a {\tt{Claim}}, {\tt{Personal Experience}}, or {\tt{Question}}. We characterize these classes as follows (we provide detailed annotation instructions in Appendix \ref{sec_apx:data_collection}): 

\vspace{0.4em}
\noindent {\tt Claim} suggests (explicitly or implicitly) a causal relationship 
    between an  \textcolor{blue}{Intervention} and an \textcolor{teal}{Outcome} (e.g., `` \textcolor{blue}{\textit{I}} completely cured my  \textcolor{teal}{\textit{O}}''). 
    Operationally, we are interested in identifying statements that might reasonably be interpreted by the reader as implying a causal link between an intervention and outcome, as this may in turn 
    influence their perception regarding the efficacy of an intervention for a particular condition and/or outcome (i.e., relationship between an \textcolor{blue}{\textit{I}} and \textcolor{teal}{\textit{O}}).

\vspace{0.4em}
\noindent {\tt Question} poses a direct question, e.g., 
    ``Is this normal?''; ``Should I increase my dosage?''.
    
\vspace{0.4em}
\noindent {\tt Personal Experience} describes an individual's experience, for instance the trajectory of their condition, or experiences with specific interventions. 

\vspace{0.4em}
This is a \emph{multi-label} scheme: Spans can (and often do) belong to more than one of the above categories. 
For example, personal experiences can often be read as implying a causal relationship.
Consider this example: ``My doctor put me on \textcolor{blue}{\emph{I}} for my \textcolor{orange}{\emph{P}}, and I am no longer experiencing \textcolor{teal}{\emph{O}}''.
This describes an individual treatment history, but could also be read as implying that \textcolor{blue}{\emph{I}} is a viable treatment for \textcolor{orange}{\emph{P}} (and specifically for the outcome \textcolor{teal}{\emph{O}}). 
Therefore, we would mark this as both a {\tt{Claim}} and a {\tt{Personal Experience}}. 
By contrast, a general statement asserting a causal relationship outside of any personal context like ``\textcolor{blue}{\emph{I}} can cure \textcolor{teal}{\emph{O}}'' is what we will refer to as a ``pure claim'', meaning it exclusively belongs to the {\tt{Claim}} category. 

In the second stage, workers are asked to further annotate ``pure claim'' instances by marking spans within them that correspond to the Populations, Interventions/Comparators,\footnote{This is the standard PICO framework, but we collapse Interventions and Comparators into the Intervention category, as the distinction is arbitrary.} Outcomes (the PIO elements) associated with the claim. 


\subsection{Crowdsourcing Annotations}
We hired crowdworkers to perform the above annotation tasks on Amazon Mechanical Turk (AMT).\footnote{We consulted with an Institutional Review Board (IRB) to confirm that this annotation work did not constitute human subjects research.} 
To estimate required annotation time and determine fair pay rates, we ran an internal pilot with two PhD students (both broadly familiar with this research area) on 100 samples.\footnote{Based on the estimate from our pilot experiments, payrate for AMT workers was fixed to US \$9 per hour for stage-1 annotations and US \$11 per hour for stage-2 annotations, irrespective of geographic location.} 
To gauge quality and recruit workers from AMT, we ran two pilot experiments in which we collected sentence-level annotations on posts sampled from three medical populations (i.e., subreddits), comprising $\sim$6,000 posts in all. 

We required 
all workers 
have an overall job approval rate of $\ge$90$\%$. 
Based on an initial set of AMT annotations we 
re-hired only workers who reliably followed annotation instructions (details in Appendix \ref{sec_apx:data_collection}), 
and we actively recruited the top workers to continue on with 
increased pay. 
We obtained annotations from at least three workers for each post, allowing for robust inference of reference labels. 
Recruited workers were also paid periodic bonuses (equivalent to two hours of pay) based on the quality of their annotated samples. 



\begin{table}[]
\centering
\begin{tabular}{@{}lc|ccc@{}}
\toprule
\multicolumn{1}{c}{} & \textbf{Fliess} $\kappa$ & \textbf{P} & \textbf{R} & \textbf{F1} \\ \midrule
Questions            & 0.86            & 0.85       & 0.82       & 0.84        \\
Claims               & 0.69            & 0.63       & 0.53       & 0.58        \\
Experiences          & 0.71            & 0.78       & 0.69       & 0.73        \\ \midrule
POP                  & 0.92            & 0.94       & 0.91       & 0.92        \\
INT                  & 0.74            & 0.76       & 0.70       & 0.73        \\
OUT                  & 0.78            & 0.73       & 0.68       & 0.70        \\ \bottomrule
\end{tabular}
\caption{Token-wise label agreement among experts measured by Fleiss $\kappa$ on a subset of data. We further compute precision, recall, and F1 scores for ``aggregated'' labels by evaluating them against unioned ``in-house'' expert labels. }
\label{tab:qval1}
\end{table}

\begin{figure*}
\centering
  \includegraphics[scale=0.37]{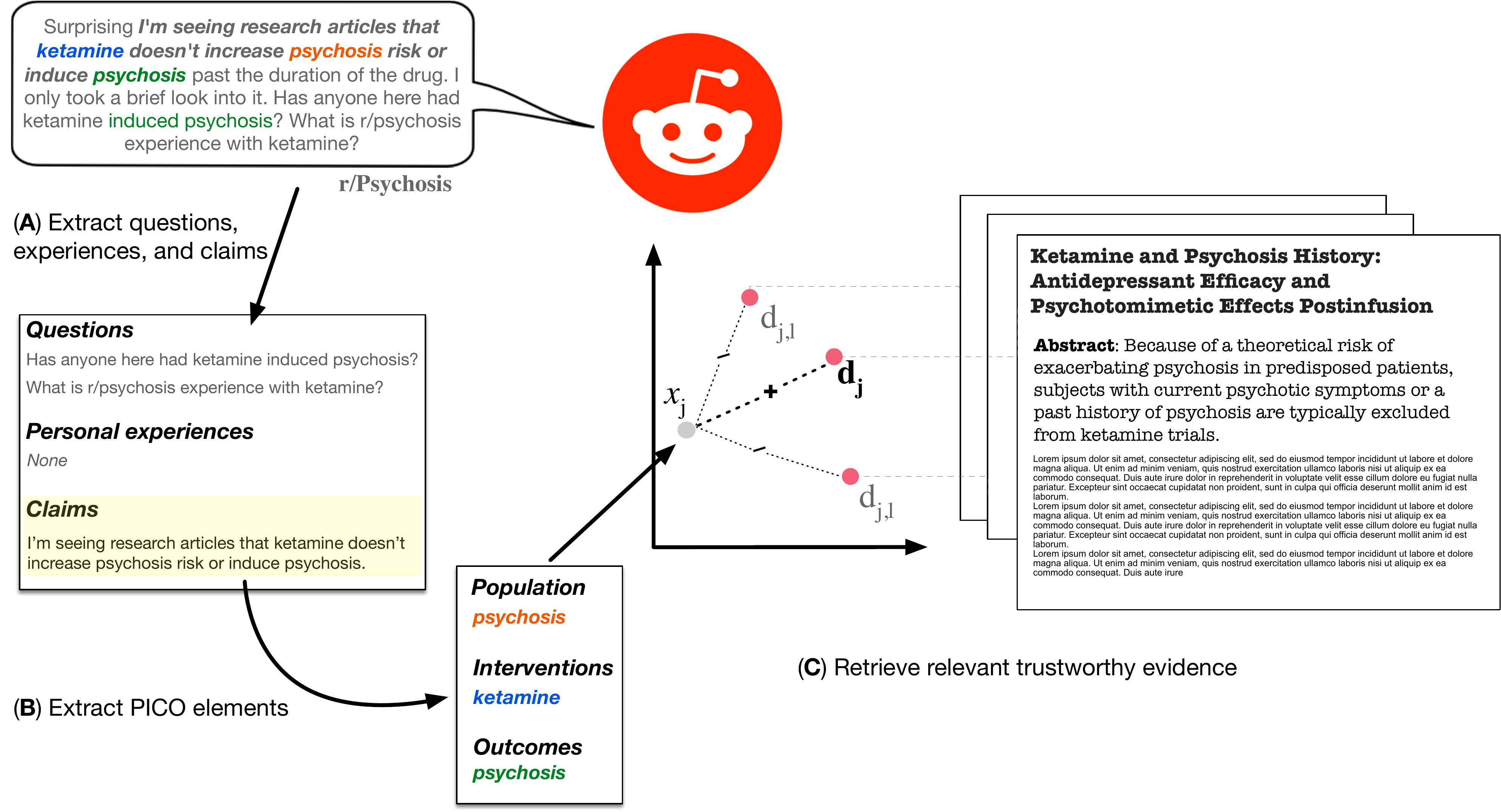}
  \caption{Examples portraying potential use cases of our corpus. We showcase three distinct tasks, to be performed in sequence.  
  The first (A) entails 
  extracting spans corresponding to claims (highlighted in \textbf{bold}) from 
  a given Reddit post. 
  The second step (B) is to identify the PICO elements associated with each claim. 
  In the final step (C), we use the outputs of the first two models with the original post to obtain a dense representation, 
  enabling us to retrieve relevant evidence from a large dataset of trusted medical evidence (e.g., PubMed).}
\end{figure*}

\subsection{Quality Validation}
\label{section:quality_val}
To evaluate annotation quality we calculate token-wise label agreement between annotators, and amongst ourselves. We emphasize here that token-level $\kappa$ for sequences is quite strict and disagreements often reflect \textit{where} annotators decide to mark span boundaries. 
Despite this, for the first stage agreement (Fleiss $\kappa$) on labeled questions, experiences, and claims was $0.62$, and for the second stage 
$0.55$. 
We consider this \textit{moderately strong} agreement, in line with agreement reported for related annotation tasks in the literature \cite{nye-etal-2018-corpus,Delger2012BuildingGS}. 
To quantify this and further gauge the quality of collected annotations, we run a few additional analyses.

As previously stated, prior to collecting annotations on Amazon MTurk, we (the authors) 
annotated a subset of data (100 samples/stage) internally to assess task difficulty and to estimate the time required for annotation. 
As an additional quality check, we use these annotations to calculate token-wise label agreement. 
Table \ref{tab:qval1} reports the results; while there remains some discrepancy owing to the inherent complexity of the task, there is higher agreement between the us than between workers.

Each of these 
samples was also annotated by three workers.
We aggregate these labels using majority-vote and compute token-wise precision-recall of these aggregated labels against the reference ``in-house'' labels (Table \ref{tab:qval1}).
We report the same metrics per annotator evaluated against aggregated MTurk labels in Table \ref{tab:qval3} (Appendix \ref{sec_apx:data_sum}). 
Despite moderate agreement between annotators, aggregated labels agree comparatively well with the ``expert'' consensus, 
indicating that while individual worker annotations are somewhat noisy, aggregated annotations are reasonably robust. 


\subsection{Dataset Details}
\label{section:dataset-summary}

Table \ref{tab:examples} provides illustrative samples from  \textbf{\textcolor{red}{Red}HOT} and Table \ref{tab:stats} provides some descriptive statistics along with examples of included health populations.
We broadly characterize populations (conditions) as \emph{Very Common}, \emph{Common} or \emph{Rare}, and sought a mix of these.
This was not the only attribute that informed which conditions we selected for inclusion in our dataset, however.
For example, we wanted a mix of populations with respect to volume of online activity (e.g., the Diabetes subreddit has over $60k$ active visitors; Lupus has $8k$). 
We also wanted to include both chronic and treatable conditions (e.g., Narcolepsy is a rare and chronic condition, while Gout is common and treatable), and mental and physical disorders (e.g., ADHD, Rheumatoid Arthritis).
Another consideration was whether a condition can be self-diagnosed or requires professional assessment (e.g., Bulimia is usually self-diagnosable but can potentially be life-threatening; Gastroparesis is chronic but requires a professional medical diagnosis).


The number of \textit{claims} across different categories of health populations are far outnumbered by \textit{questions} ($\sim$10x) and \textit{experiences} ($\sim$13x). 
The average post length is $\sim$117 tokens while the average length of a \textit{claim} within a post is $\sim$20 tokens. 
\textit{Questions} and \textit{experiences} have average lengths of $\sim$11 and $\sim$27 tokens, respectively. 
We provide per condition statistics in Appendix \ref{sec_apx:data_sum}. 

\section{Tasks and Evaluation}
\label{section:tasks}

\textbf{\textcolor{red}{Red}HOT} may support a range of tasks related to processing health-related social media posts.
Here we focus on an 
important, timely task: Identifying
medical claims on social media, and then retrieving relevant and trustworthy evidence that may support or refute them. 
Methods for this task could aid content moderation on health-related forums, by providing an efficient means to (in)validate claims. 
More generally, such methods may permit meaningful ``fact checking'' of health-related claims by providing relevant contextualizing evidence.

We outline 
a three-step approach 
for this task. 
(1) Identify spans/sentences corresponding to \textit{pure} claims. 
(2) Extract from these specific PICO elements.
(3) Retrieve clinical literature --- specifically, reports of RCTs --- relevant to the claim, i.e., the extracted PIO elements. 
We limit our focus to the problem of evidence retrieval here; future work might consider the subsequent step of automated claim validation on the basis on this. 

Below we assess components for each of these steps. 
For the span and PIO extraction steps (1 and 2), we evaluate models retrospectively under standard classification metrics (i.e. precision, recall, and F1 scores) using fixed train, development, and test sets which we will distribute with \textbf{\textcolor{red}{Red}HOT}. 
The final step (3) requires relevance judgments to evaluate model performance; for this we enlisted medical doctors (Section \ref{sec:human_eval}). 

\subsection{Identifying Claims, Experiences, Questions, and PIO Elements}

We treat the first two steps as sequence tagging tasks for which we evaluate two types of models: A simple linear-chain Conditional Random Field (CRF; \citealt{crf}), and Transformer-based models \cite{vaswani2017attention} --- specifically BERT variants \cite{devlin-etal-2019-bert,Liu2019RoBERTaAR}.\footnote{We also explored t5 \cite{JMLR:v21:20-074} with middling results, which we report in the Appendix.}
The features for the CRF we use are: Indicators of next, previous, and current words; Part-of-speech tags,\footnote{Extracted with SciSpacy (\url{https://allenai.github.io/scispacy/}).} and; Indicators encoding if sentences contain digits, uppercase letters, and/or measurement units. 
BERT variants yield contextualized representations of input tokens, 
which we then use to predict labels (i.e., Claims, Experiences and Questions) by adding a linear layer on top of the encoder outputs. PIO elements are extracted using a concatenated input of the original Reddit post and an identified claim.

\begin{table*}[]
\small
\centering
\begin{tabular}{@{}lcccccc@{}}
\toprule
              & P     & R     & F1                        & F1$_{\text{POP}}$ & F1$_{\text{INT}}$ & F1$_{\text{OUT}}$ \\ \midrule
\texttt{BERT} \cite{devlin-etal-2019-bert}         & 43.88 & 36.13 & \multicolumn{1}{c|}{39.62} & 41.77   & 44.68   & 33.05   \\
\texttt{BioRedditBERT} \cite{basaldella-etal-2020-cometa} & 44.44 & 36.55 & \multicolumn{1}{c|}{40.12} & 41.92   & 44.31   & 34.61   \\
\texttt{biomedRoBERTa} \cite{domains} & 38.80 & 21.48 & \multicolumn{1}{c|}{27.66} & 30.54   & 28.13   & 24.54   \\
\texttt{RoBERTa} \cite{Liu2019RoBERTaAR}      & \textbf{47.45} & \textbf{39.27} & \multicolumn{1}{c|}{\textbf{42.97}} & \textbf{46.09}   & \textbf{45.99}   & \textbf{36.38} \\
t5-small \cite{JMLR:v21:20-074} & 41.49 & 38.55 & \multicolumn{1}{c|}{39.97} & 39.61   & 45.02   & 32.41  \\ \bottomrule
\end{tabular}
\caption{Results on the test set for the token-level PICO tagging task. }
\label{tab:res-pico}
\end{table*}

\begin{table*}[]
\small
\centering
\resizebox{16cm}{!}{
\begin{tabular}{@{}lccccccccc@{}}
\toprule
\multicolumn{1}{c}{} & \multicolumn{3}{c}{Claims}                                                                 & \multicolumn{3}{c}{Experiences}                                                   & \multicolumn{3}{c}{Questions}                                                     \\ \cmidrule(l){2-10} 
\multicolumn{1}{c}{} & F1                        & P                         & R                                  & F1                        & P                         & R                         & F1                        & P                         & R                         \\ \midrule
CRF \cite{crf}                  & 33.87                     & 35.61                     & 32.29                              & 40.08                     & 40.52                     & 39.64                     & 86.89                    & 85.55                     & 88.27    \\
\texttt{BERT} \cite{devlin-etal-2019-bert}                & \textbf{52.63}            & 58.82                     & 47.61                              & 56.68                     & 59.46                     & 54.33                     & 92.39                     & 88.76                     & 96.34 \\
\texttt{RoBERTa}    \cite{Liu2019RoBERTaAR}          & \multicolumn{1}{l}{47.05} & \multicolumn{1}{l}{61.53} & \multicolumn{1}{l}{\textbf{38.09}} & \multicolumn{1}{l}{56.81} & \multicolumn{1}{l}{57.11} & \multicolumn{1}{l}{56.52} & \multicolumn{1}{l}{93.06} & \multicolumn{1}{l}{89.01} & \multicolumn{1}{l}{98.34} \\
\texttt{BioRedditBERT}  \cite{basaldella-etal-2020-cometa}      & 45.16                     & \textbf{70.92}            & 33.29                              & \textbf{59.51}            & \textbf{62.49}            & \textbf{58.92}            & \textbf{93.61}            & \textbf{89.29}            & \textbf{98.37}            \\ \bottomrule
\end{tabular}
}
\caption{Results on the test-set of span-classification to identify pure claims, questions, and experiences.}
\label{tab:task1_res}
\end{table*}

\subsection{Evidence Retrieval}
\label{sec:evi_retrieval_description}

For the retrieval task 
we assume the model is given: (i) The original Reddit post and a claim; (ii) PIO elements associated with that claim, and; (iii) A \textit{large} set of candidate articles featuring trustworthy evidence to rank.
We use $\sim$800,000 abstracts from \emph{Trialstreamer}\footnote{\url{https://trialstreamer.ieai.robotreviewer.net/}} \cite{10.1093/jamia/ocaa163}, a continuously updated database of reports of randomized controlled trials (RCTs). 
RCTs are appropriate here because of our focus on causal claims --- results from randomized trials are the most reliable means of evaluating such assertions \cite{MELDRUM2000745}.


\subsubsection{Task Formulation}

Formally, we represent a single input instance as $(p
, c_j, {\text {pop}}_j, {\text{int}}_j, {\text{out}}_j)$ where $p$ is a post comprising $n$ sentences, 
$c_j$ is the $j$th claim, and ${\text{pop}}_j, {\text {int}}_j, {\text{out}}_j$ are the \emph{sets} of populations, interventions, outcomes associated with claim $j$. 

The model is tasked with finding relevant 
abstracts from the candidate set $\mathcal{A}$,
which comprises 
abstracts from published clinical trial reports. 
This is particularly challenging because a large number of candidates can mention the same set of PIO entities (i.e., investigate the same interventions and/or outcomes), but in a context 
unrelated to the 
claim being made in the social media post. 
This may be especially problematic for retrieval methods based primarily on string overlap measures. 
We therefore propose a learning based approach.
This requires supervision; we next describe our approach to deriving this automatically.

\subsubsection{Pseudo Training Data}
\label{sec:pseudo_data}
Supervised 
neural retrieval models require annotations indicating the relevance of instances (here, published evidence) to inputs (claims on social media). 
We do not have such judgments, and so instead derive ``pseudo'' training data automatically.

We started with $\sim$800,000 
abstracts of medical RCTs in Trialstreamer.
We then used  
Reddit posts containing pure claims as \emph{templates} to create pseudo matches between 
medical claims and abstracts. 
Specifically, we substituted annotated PIO elements in claims made within Reddit posts with PIO elements sampled from Trialstreamer abstracts. (Trialstreamer includes PICO elements automatically extracted from all articles that it indexes.)
This yields pairs of (a) naturally occurring claims (with their PIO spans replaced) and (b) RCT abstracts that are relevant to said claims by construction. 
We provide examples of this pseudo matching in Appendix \ref{sec_apx:examples}. 
We generated a total of 85,000 examples of \texttt{(pseudo claims, evidence abstract)} one-to-many pairs 
to be used to train a neural retrieval model (described below). 
The generated examples may be noisy, but hopefully sufficient to train a model to retrieve medical abstracts relevant to health-related claims made on social media.




\begin{table*}[]
\centering
\resizebox{15cm}{!}{
\begin{tabular}{@{}lcccccccccc@{}}
\toprule
                                                                  & \multicolumn{5}{c}{\textbf{MRR @k}}                                                                     & \multicolumn{5}{c}{\textbf{Precision @k}}                                           \\ \cmidrule(l){2-11} 
\multicolumn{1}{c}{\textbf{k}}                                  & 1              & 5              & 10             & 50             & \multicolumn{1}{c|}{100}            & 1              & 5              & 10             & 50             & 100            \\ \midrule
random                                                            & 0.00           & 0.003          & 0.02           & 0.02           & \multicolumn{1}{c|}{0.02}           & 0.00           & 0.02           & 0.00           & 1.10           & 2.80           \\
BM25                                                              & 5.34           & 7.98           & 9.86           & 14.36          & \multicolumn{1}{c|}{16.70}          & 5.34           & 10.40          & 14.45          & 26.20          & 33.14          \\
DPR \cite{karpukhin-etal-2020-dense}                                                              & 8.07           & 10.96          & 11.89          & 12.20          & \multicolumn{1}{c|}{13.77}          & 8.07           & 16.50          & 23.58          & 31.98          & 36.87          \\ \midrule
\multicolumn{10}{c}{\small{\textit{(trained on the \textbf{\textcolor{red}{Red}HOT} pseudo training set)}}}                                                                                                                                                                           \\
\textbf{\textcolor{red}{Red}HOT}-DER (\texttt{BERT}-based)    & 39.14          & 47.99          & 49.3           & 50.28          & \multicolumn{1}{c|}{50.35}          & 39.14          & 62.55          & 72.64          & 83.73          & 91.74          \\
\textbf{\textcolor{red}{Red}HOT}-DER (\texttt{RoBERTa}-based) & \textbf{45.93} & \textbf{54.60} & \textbf{55.90} & \textbf{56.73} & \multicolumn{1}{c|}{\textbf{56.78}} & \textbf{45.93} & \textbf{69.90} & \textbf{78.81} & \textbf{94.73} & \textbf{98.06} \\ \bottomrule
\end{tabular}
}
\caption{Results of evidence retrieval baselines evaluated on pseudo test data.}
\label{tab:retrieval-results}
\end{table*}

\subsubsection{\textcolor{red}{Red}HOT Dense Evidence Retriever} 

We 
train a neural retrieval model 
on the \textbf{\textcolor{red}{Red}HOT} corpus, using a setup similar to DPR \cite{karpukhin-etal-2020-dense}. 
We first assemble a collection of $m$ 
RCT abstracts to create an evidence corpus, ${\mathcal{A}=\{d_1, d_2,...,d_{m}\}}$. 
There are hundreds of thousands of RCTs, so we need an \textit{efficient} retriever that can select a small set of relevant abstracts. 
Formally, a retrieval operation $\{R$: $(x_j,\mathcal{A})\rightarrow\mathcal{A_F}\}$ accepts an input contextualizing string $x_j$ 
and a corpus of evidence $\mathcal{A}$, and returns a \textit{much smaller} filtered set $\mathcal{A_F}\subset\mathcal{A}$, where $|\mathcal{A_F}| = k$. 

We form an input context string $x_j$ for a claim $j$ made within a post $p$ by concatenating the post, claim, and PIO elements extracted from the claim: $x_j = [p \oplus c_j \oplus \text{pop}_j \oplus \text{int}_j \oplus \text{out}_j]$,
 where $\oplus$ denotes concatenation with {\tt [SEP]} tokens.
%
We define two dense neural encoders ($E_C$, $E_D$; both initialized with RoBERTa-base) to project the context string $x_j$, and evidence (abstracts) from $\mathcal{A}$ to fixed 768 dimensional vectors. 
Similarity between the context string and evidence abstract is defined using the dot product of their vectors, $\phi(x_j,d_l) = E_{C}(x_j)^{T} E_{D}(d_l).$ 

We train the model to minimize 
the negative log-likelihood of the positive evidence such that it pushes the context string vector $x_j$ close to 
the representation of relevant evidence $d_{j}^{+}$, and away from $b$ irrelevent abstracts ($d^{-}_{j1}, d^{-}_{j2}, ... d^{-}_{jb}$) in the same mini-batch\footnote{We set the size of the mini-batch to 100.} (``in-batch negative sampling''):
\begin{equation*}
    \mathcal{L} = 
    \frac{\exp \phi(x_j, d_{j}^{+})}{\exp \phi(x_j, d_{j}^{+}) + \sum_{l=1}^{b}\exp \phi(x_j, d^{-}_{jl})}
\end{equation*}

\noindent In-batch negative sampling 
has been shown to be effective for dual-encoder training \cite{DBLP:journals/corr/HendersonASSLGK17, gillick-etal-2019-learning}. 
Here, all samples in a minibatch are taken from the same \textit{population} (condition) set, e.g., a mini-batch with a sample containing a claim about \texttt{diabetes} will have negative evidence abstracts that are also related to \texttt{diabetes}. 

For test examples, we rank all evidence (abstracts in Trialstreamer) according to their similarity to the context string.
To do this efficiently, we induce representations of all 
the abstracts in the Trialstreamer database using the evidence encoder and index these using the Facebook AI Similarity Search library \cite{8733051}.\footnote{FAISS: Open-source library for efficient similarity search and clustering of dense vectors; \url{https://ai.facebook.com/tools/faiss/}.}



\subsubsection{Baseline Models}


\paragraph{\textbf{BM25}} A standard Bag-of-Words method for IR~\cite{robertson1995okapi}. 
We form queries by concatenating the Reddit post with a single claim and its corresponding PIO frames. We used a publicly available BM25 implementation from the Rank-BM25 library.\footnote{\url{https://github.com/dorianbrown/rank_bm25}} 

\paragraph{Dense Passage Retrieval (\textbf{DPR})} is a dense retrieval model 
trained to retrieve \textit{relevant} context spans (``paragraphs'') in an open domain question-answering setting~\cite{karpukhin-etal-2020-dense}.
In general, such models map \textbf{queries} and \textbf{candidates} to embeddings, 
and then rank candidates with respect to a similarity measure (e.g., dot product) taken between these. While originally designed for open-domain question answering, use of DPR-inspired models has been extended to general retrieval tasks \cite{thai2022relic}.
We use a DPR context encoder trained on Natural Questions \cite{47761} with dot product 
similarity
.\footnote{\url{https://huggingface.co/facebook/dpr-ctx-encoder-single-nq-base}}

\subsection{Results}
\label{sec:evaluation}
We evaluate models for the tasks of identifying claims, experiences, and questions and extracting PIO elements using 
precision, recall, and F1 scores.
We report results per class for the first task in Table \ref{tab:task1_res}. 
\texttt{BioRedditBERT} \cite{basaldella-etal-2020-cometa} --- a BERT model initialized from \texttt{BioBERT} \cite{10.1093/bioinformatics/btz682} and further pre-trained on health-related Reddit posts --- fares best here. We report results for the second task (PIO tagging) in Table \ref{tab:res-pico}.\footnote{Results from additional experiments using other model variants are reported in Appendix \ref{sec_apx:add_results}.}
Here \texttt{RoBERTa} \cite{Liu2019RoBERTaAR} modestly outperforms \texttt{BioRedditBERT} \cite{basaldella-etal-2020-cometa}.

Models 
for the retrieval task 
rank evidence candidates for each input (post, claim, PIO frame).
We therefore use standard ranking metrics for evaluation, including mean reciprocal rank, and precision\symbol{`@}$k$ (for $k = 1,5,10,50,100$). Baseline results are reported in Table \ref{tab:retrieval-results}. 
We emphasize that these results are with respect to pseudo annotated data, effectively providing an unfair advantage to \textbf{\textcolor{red}{Red}HOT}-DER, given that this was optimized on data from this distribution. 
We report results with respect to manual relevance judgments provided by experts in Section \ref{sec:human_eval}. 


As we might expect, the pre-trained neural DPR model outperforms the naive string matching BM25 method. 
Furthermore, as anticipated, explicitly training for evidence retrieval confers pronounced advantages: \textbf{\textcolor{red}{Red}HOT}-DER fares $\sim$8x better than BM25 
and $\sim$5x better than ``off-the-shelf'' pre-trained DPR \cite{karpukhin-etal-2020-dense} with respect to retrieving relevant evidence (precision@1) corresponding to medical claims. 
Again, this is not particularly surprising given that we are evaluating models with respect to the pseudo annotations with which \textbf{\textcolor{red}{Red}HOT}-DER was trained (because we do not otherwise have access to explicit relevance judgments). 
Therefore, we next present results from more meaningful manual relevance evaluations performed by domain experts.

\begin{table}[]
\centering
\begin{tabular}{@{}lcccc@{}}
\toprule
\multicolumn{5}{c}{\textbf{Cumulative \# of relevant abstracts @$k$}} \\ \cmidrule(l){2-5} 
\multicolumn{1}{c}{\textbf{k}}    & 1     & 3      & 5      & 10    \\ \midrule
\multicolumn{5}{c}{\footnotesize{\textit{Pre-trained DPR \cite{karpukhin-etal-2020-dense}}}}                        \\
\textbf{Relevant}                 & 6     & 16     & 29     & 58    \\
\textbf{Somewhat relevant}        & 14    & 39     & 66     & 135   \\
\textbf{Irrelevant}                & 80    & 245    & 405    & 807   \\ \midrule
\multicolumn{5}{c}{\footnotesize{\textit{\textbf{\textcolor{red}{Red}HOT}-DER trained on pseudo data}}}    \\
\textbf{Relevant}                 & 18    & 62     & 101    & 201   \\
\textbf{Somewhat relevant}        & 17    & 49     & 87     & 193   \\
\textbf{Irrelevant}               & 65    & 189    & 312    & 606   \\ \bottomrule
\end{tabular}
\caption{Results from manual (domain expert) evaluations for DPR and our pseudo-supervised DER model.}
\label{tab:eval_ret_res}
\end{table}


\subsection{Expert Manual Relevance Judgments} 
\label{sec:human_eval}
We evaluated models in terms of retrieving evidence relevant to \textit{naturally occurring} medical claims, as opposed to the \textit{pseudo} data derived for training.
We hired three domain experts (medical doctors) on the Upwork platform.\footnote{Upwork (\url{https://www.upwork.com/}) allows clients to interview, hire and work with freelancers. All of our evaluators had medical degrees and were hired at wages ranging from \$15 to \$20 per hour for a minimum of 15 hours.}  
Providing hundreds of retrieved medical abstracts per claim to a human evaluator for assessment is infeasible, so we instead provided evaluators with 10 retrieved abstracts each 
for 100 
individual claims, retrieved using the pretrained DPR \cite{karpukhin-etal-2020-dense} model and our \textbf{\textcolor{red}{Red}HOT}-DER trained on pseudo data. 
(We compared the proposed distantly supervised model to DPR because it is the strongest baseline we evaluated in preliminary experiments.)

We asked evaluators to 
categorize each retrieved abstract as: (1) \texttt{Relevant}; (2) \texttt{Somewhat Relevant}, or; (3) \texttt{Irrelevant} to the corresponding claim. 
An abstract was to be considered \texttt{Relevant} if and only if it (1) contained to the same \textbf{\textcolor{orange}{P}}, \textbf{\textcolor{blue}{I}}, and \textbf{\textcolor{teal}{O}}
elements mentioned in the original Reddit post, \textbf{and} (2) provided information to support or refute the claim in question. 
An abstract might be deemed \texttt{Somewhat Relevant} if it contains a 
\textbf{\textcolor{orange}{P}}, \textbf{\textcolor{blue}{I}}, and \textbf{\textcolor{teal}{O}} set in line with the given claim, but does not provide any information relating these elements. 
We provide examples in the Appendix \ref{sec_apx:examples}.

Human evaluators achieve strong agreement: All three evaluators chose the same relevance label 71.33\% of the time, while they all chose a different label only in 1.29\% of the total instances. 
They also show substantial agreement in terms of Fleiss $\kappa$ (0.71).
We derive final relevance labels by majority vote. 
Comparing results from Table \ref{tab:retrieval-results} and Table \ref{tab:eval_ret_res}, at $k=1$ we see similar values of precision in the 
manually annotated data 
and pseudo test data. 
However, for higher values of $k$ large differences emerge,  indicating considerable room for improvement. 
Compared to the pre-trained DPR model, at $k=1$ \textbf{\textcolor{red}{Red}HOT}-DER retrieves a substantially larger fraction of relevant evidence abstracts (3x). 
At higher $k$, we also observe a large reduction in the number of \textit{irrelevant} abstracts retrieved (e.g., at $k=10$, the number of irrelevant abstracts decreases by $\sim30\%$).
We believe this highlights the value of our proposed distant supervision scheme.


\section{Related Work}
\label{section:related-work}
\vspace{0.4em}
\noindent{\bf Claim validation via evidence retrieval}
\label{sec:rel_work_claims}
Past work has typically treated (open domain) claim validation as a two-step process in which one retrieves evidence relevant to a given claim, and then makes a prediction regarding claim validity on the basis of this. 
Information retrieval (IR) models are usually used in the first step to rank order documents based on relevance to a given claim \cite{thorne-etal-2018-fever, wadden-etal-2020-fact, thai-etal-2022-relic, hanselowski-etal-2018-ukp, samarinas-etal-2021-improving, saeed-etal-2021-neural}. 
The next step is usually to characterize \textit{retrieved} evidence as {\tt supporting}, {\tt refuting}, or {\tt not providing enough information} (although this latter category is not always included). Evidence might be individually characterized \cite{pradeep-etal-2021-scientific}, or aggregated to make a single prediction about the veracity of the claim \cite{sarrouti-etal-2021-evidence-based}.

\vspace{0.4em}
\noindent{\bf Scientific claim verification} Beyond ``general domain'' verification, there have been efforts focused specifically on vetting \emph{scientific} claims.
{\tt SciFact} \cite{wadden-etal-2020-fact} largely follows the typical fact verification setup outlined above (but for scientific claims).
Subsequent efforts have focused specifically on verifying claims related to COVID-19 \cite{saakyan2021covid}. 
The {\tt evidence inference} task \cite{lehman-etal-2019-inferring,deyoung2020evidence} entails inferring whether a given trial report supports a significant effect concerning a specific intervention, comparator, and outcome. 

\vspace{0.4em}
\noindent{\bf Crowd-sourcing annotation of scientific and medical texts}
We have relied on crowdworkers to annotate the instances comprising \textbf{\textcolor{red}{Red}HOT}. 
This is in keeping with a body of work that has shown crowdworkers capable of annotating health-related texts, even when these are technical \cite{drutsa-etal-2021-crowdsourcing}. 
For example, several past efforts have crowdsourced annotation of texts drawn from PubMed, e.g. for mentions of diseases \cite{nye-etal-2018-corpus,good2014microtask}. More recently, \citet{bogensperger-etal-2021-dreamdrug} crowdsourced a dataset of drug mentions (a type of \textit{intervention}) on the darknet. \citet{khetan-etal-2022-mimicause} crowdsourced annotations of electronic health records to identify causal relations between medical entities. Similarly, there is a body of work relying on crowdsourcing to accomplish a diverse set of domain-specific non-medical NLP tasks \cite{sukhareva-etal-2016-crowdsourcing, fromreide-etal-2014-crowdsourcing, bhardwaj-etal-2019-carb, gardner-etal-2020-determining}. 

\vspace{0.4em}
\noindent{\bf Health-related Reddit corpora} Past work has also built corpora of health-related Reddit posts. For example, \citet{cohan-etal-2018-smhd} assembled a dataset of Reddit posts made by individuals who self-reported one of nine mental health diagnoses of interest. 
Building on this work, \citet{jiang-etal-2020-detection} introduced a dataset of Reddit posts to evaluate models for automatically detecting psychiatric disorders.

\section{Conclusions}
\label{sec:conclusions}
We presented \textbf{\textcolor{red}{Red}HOT}: a new, publicly available dataset comprising of about 22,000 richly annotated Reddit posts extracted from 24 medical condition-based communities (``subreddits''). 
This dataset 
meets a need for corpora that can facilitate development of language technologies for processing health-related social media posts. 

We evaluated baseline models for categorizing posts as containing claims, personal experiences, and/or questions.
Focusing on claims, we then proposed and evaluated models for extracting descriptions of populations, interventions, and outcomes, and then using such snippets to inform retrieval of trustworthy (published) evidence relevant to a given claim. 
To this end, we introduced a heuristic supervision strategy, and found that this outperformed pre-trained retrieval models. 

\section*{Limitations}


We have introduced a new annotated dataset of medical questions, experiences, and claims across a range of health populations from social media. 
We showed that this data can be used to train models potentially useful for downstream applications, e.g., by facilitating content moderation. 
However, there are important limitations to this work, specifically with respect to the raw data we sampled and the annotations on this that we have collected. 

First, the dataset we have annotated is inherently limited. 
While we have tried to select a diverse set of health populations (i.e., subreddits), these nonetheless constitute a small sample of the diverse set of existing health conditions.
Moreover, our selection has led to a corpus comprising nearly entirely of English-language posts, which is a clear limitation. 

We relied on non-expert (layperson) workers from Amazon Mechanical Turk (AMT) to carry out the bulk of annotation work. 
While we 
took steps to try and ensure annotation quality (described in Section \ref{section:data}), 
we nonetheless acknowledge that these annotations will contain noise. 
This is especially true given that 
AMT workers are not medical-experts and ultimately do not have (nor are they expected to have) sufficient knowledge of different kinds of medical terms appearing in the dataset (e.g., SSRIs' stand for \textit{selective serotonin reuptake inhibitor} and is a common form of intervention which may lead to outcomes like dizziness, anxiety, and/or insomnia, but many laypeople might simply be unaware of ordinary meaning of complicated medical terms leading them to \textit{not} matching all or part of such terms to their respective labels). 

In Section \ref{sec:pseudo_data}, we describe how we obtained \textit{pseudo} training labels to build a supervised dense retriever. To generate this data, several natural language claims get reused with substitute set of populations/interventions/outcomes. This heuristic may 
induce certain biases (as evident from Table \ref{tab:eval_ret_res} and Table \ref{tab:retrieval-results}). 
An ideal way to train a dense retriever here would be to collect positive annotation labels for \textit{every} claim in our dataset.
Collecting such supervision at scale sufficient for model training would be expensive, given that one would strongly prefer expert (medical doctor) annotations concerning the factual accuracy of claims.



\section*{Ethics Statement}

This work has the potential to contribute to human well-being by supporting development of language technologies for processing health-related social media posts. 
Such models might in turn provide insights about patient experiences and viewpoints in general, and more specifically may help community moderators identify and remove posts containing medical misinformation.

Realizing these potentially positive contributions requires annotated data with which to train relevant models; such data is the main contribution on offer in this work. 
However, releasing an annotated corpus of health-related social media posts raises concerns regarding individual privacy. 
The Reddit posts we have assembled and collected annotations were posted publicly on the Internet (almost always under pseudonyms), but nonetheless we have taken steps to ensure that individuals can choose not to be represented in this dataset. 

Specifically, we sent a message to every user in the \textbf{\textcolor{red}{Red}HOT} explaining our intent to construct and release this dataset and offering the option to ``opt out''. 
In addition, although this is not required by Reddit, we have decided not to release the collected \textit{posts} directly.  Instead we release a script that will download the posts comprising our data on-demand and align these with the collected annotations.
This means that if a user chooses to delete their post(s) from Reddit, they will also effectively be removed from our dataset. Further, we require anyone accessing this data to self-certify that they have obtain prior approval from their own IRB concerning the use-cases of their research.

\section*{Acknowledgements}

This work was supported in part by the National Science Foundation (NSF) CAREER award 1750978. 

\bibliography{anthology,custom}
\bibliographystyle{acl_natbib}

\clearpage

\appendix
\section*{Appendix for ``\textcolor{red}{Red}HOT: A Corpus of Annotated Medical Questions, Experiences, and Claims on Social Media''}

\begin{figure*}
\centering
  \includegraphics[scale=0.23]{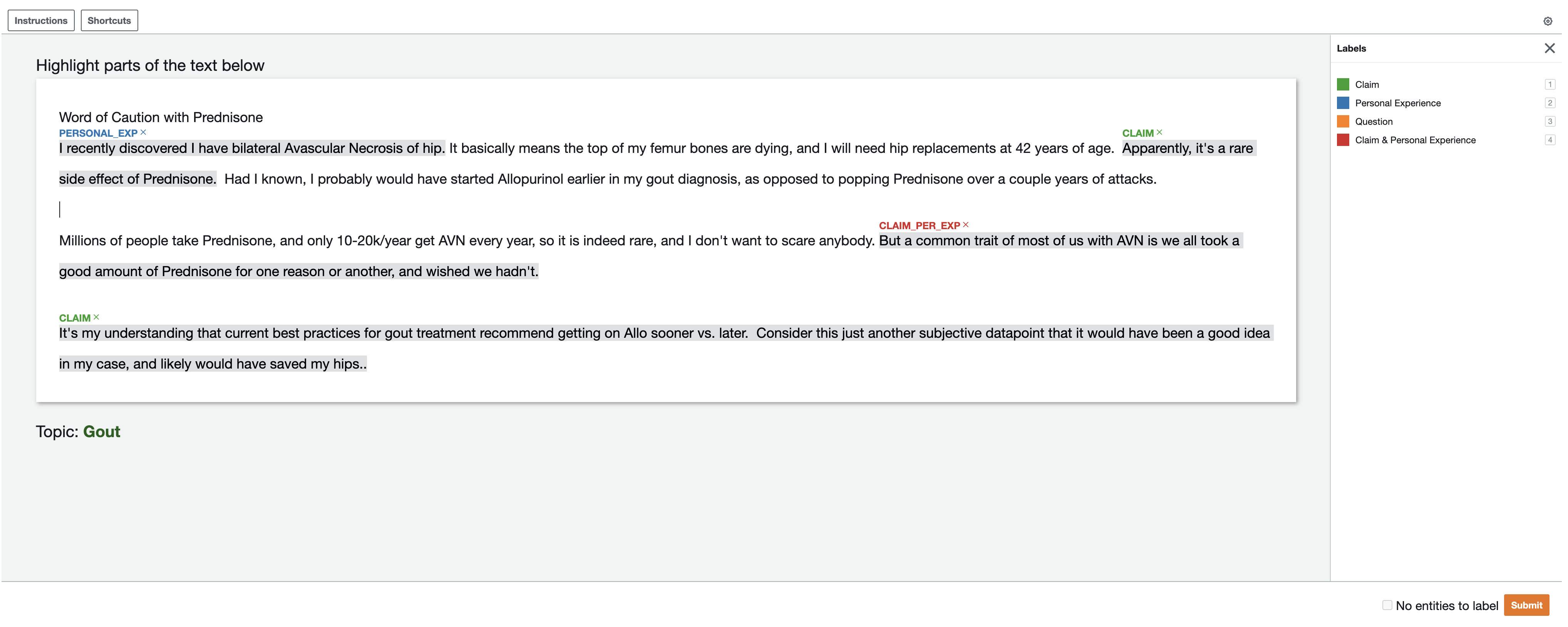}
  \caption{Stage-1 annotations interface for demarcation of spans associated with \textit{questions}, \textit{experiences}, and \textit{claims}. }
  \label{fig_apx:stg1_outlay}
\end{figure*}

\begin{figure*}
\centering
  \includegraphics[scale=0.23]{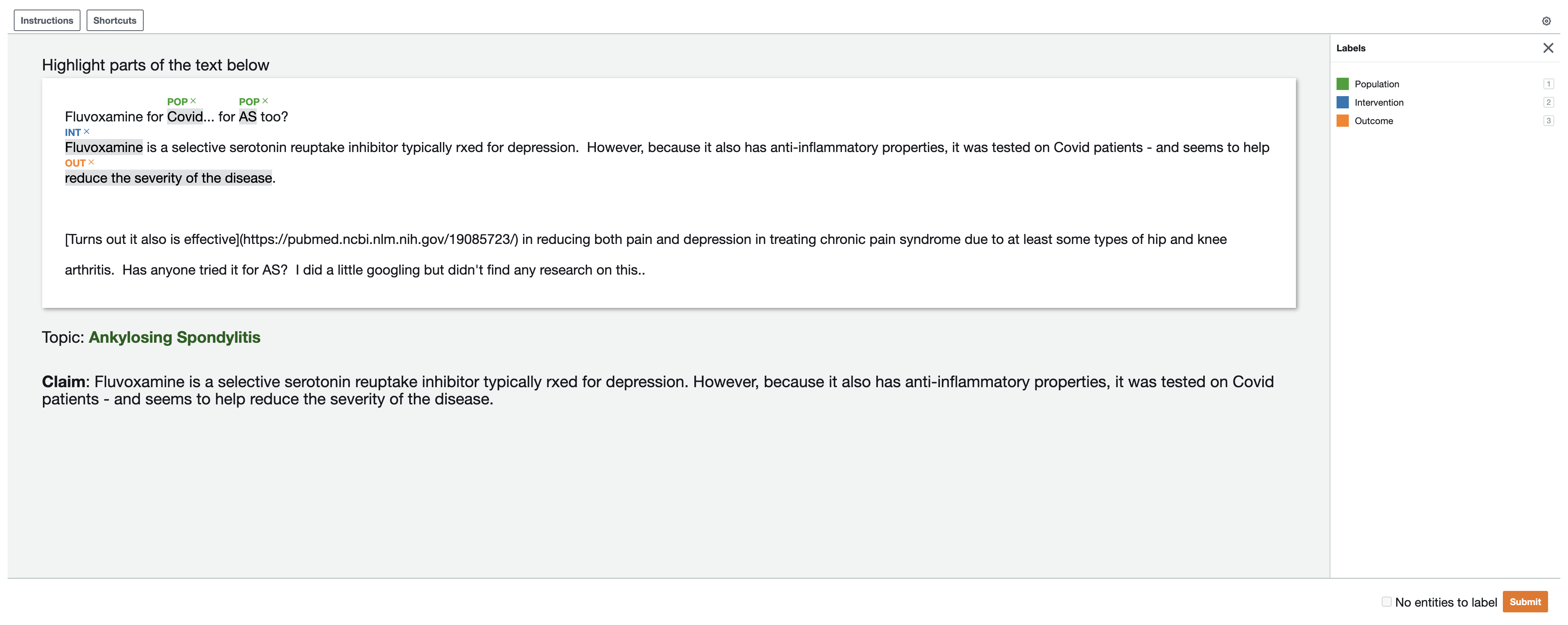}
  \caption{Stage-2 annotations interface for demarcation of PICO frames associated with a given Reddit post and a claim.}
  \label{fig_apx:stg2_outlay}
\end{figure*}

\section{Data Collection}
\label{sec_apx:data_collection}
\subsection{Sampling from Reddit}
We retrieved the \textit{newest} 1,000 posts from the respective subreddits using the Reddit \texttt{PRAW}\footnote{https://github.com/praw-dev/praw} API. 
While we could have relied on alternative sampling strategies --- e.g., ranking posts according to ``hot'' or ``best'' under Reddit's metrics --- retrieving the newest posts yields an unfiltered snapshot of the full variety of posts made to social media. 
We also considered performing completely uniform sampling over all posts ever made to a given forum, but the Reddit API limits callers to retrieving 1000 posts for any search criteria; this practically precludes uniform sampling across all time periods.
\paragraph{Preprocessing} We identified and removed all non-English text post extraction.\footnote{Langid is a python tool that allows filtering data by language: \url{https://github.com/saffsd/langid.py}.} Reddit allows its users to post media content (images/videos) in addition to text, and such imagery can be explicit or disturbing. 
Therefore, we only retained posts that did not contain any media content.  

\subsection{Annotations on Amazon Mechanical Turk}
Amazon Mechanical Turk (AMT) is a popular platform for recruiting non-expert workers to perform ``micro-tasks'' (here, annotation). 
We initially recruited workers by collecting annotations on relatively simple examples for which we already had ground truth labels. 
We provided AMT workers with a comprehensive set of instructions including (templated) examples of the respective categories. For instance: 
\begin{itemize}
    \item \textbf{Questions:} Does this work?; Are X, Y, Z symptoms normal for Condition A?; Will increasing my dosage of X help Y in any way?
    \item \textbf{Personal Experiences:} I was diagnosed with X and have since experienced symptoms Y,Z.; I took X and it seemed to help.; My mother took Y and it helped improved her Z
    \item \textbf{Claim:} My doctor told me that X should help with Y; Since increasing dosage of Z, my X levels have normalized (also an example of personal-experience); I heard from multiple people that A helps with C; I read online that X $\&$ Y are directly causing Z; I heard from my cousin that X helps control Z
\end{itemize}

For additional context we provided workers with the ``Topic'', i.e., the subreddit from which the post being annotated was sampled. 
For example, if the topic was ``Diabetes'', the piece of text will (presumably) be about diabetes, its treatments, individual experiences with the condition, and so on. 
We highlight the stage-1 annotation interface in Figure \ref{fig_apx:stg1_outlay}. 
The complete set of instructions we provided to AMT workers are available at \url{https://anonymous.4open.science/r/med_val-64C2/stg1_instructions.pdf}. 

We retained all qualified AMT workers from stage-1 to carry out additional annotations for us in stage-2, with a higher pay rate. 
The objective here was to recruit people who had established a working understanding of the data, and would presumably be proficient as a result.
Similar to stage-1, we provided workers with a comprehensive set of instructions containing (templated) examples to give a sense of what might be qualify as PIO elements:
\begin{itemize}
    \item \textbf{Populations} coronavirus, asthma, narcoleptic, diabetic, children, young, women etc.
    \item \textbf{Interventions} diet, aspirin, allopurinol, insulin, exercise, botox etc
    \item \textbf{Outcomes} depression, sweating, anxiety, pain, flares, covid etc
\end{itemize}
Interface used for stage-2 annotations is provided in Figure \ref{fig_apx:stg2_outlay}. Complete set of stage-2 instructions provided to AMT workers are available at \url{https://anonymous.4open.science/r/med_val-64C2/stg2_instructions.pdf}. 

\section{Dataset Summary}
\label{sec_apx:data_sum}

\begin{table*}[]
\centering
\resizebox{16cm}{!}{
\begin{tabular}{@{}lrrrrrr@{}}
\toprule
\multicolumn{1}{c}{\textbf{\begin{tabular}[c]{@{}c@{}}Population Type\\ (subreddit)\end{tabular}}} & \multicolumn{1}{c}{\textbf{\begin{tabular}[c]{@{}c@{}}\# of posts in\\ \textcolor{red}{Red}HOT\end{tabular}}} & \multicolumn{1}{c}{\textbf{\begin{tabular}[c]{@{}c@{}}Avg. length of \\ post (\# tokens)\end{tabular}}} & \multicolumn{1}{c}{\textbf{\# claims}} & \multicolumn{1}{c}{\textbf{\# questions}} & \multicolumn{1}{l}{\textbf{\# experiences}} & \multicolumn{1}{c}{\textbf{\begin{tabular}[c]{@{}c@{}}\# subscribers\\ on Reddit\end{tabular}}} \\ \midrule
Dysthymia                                                                      & 999                                                                                          & 175.42                                                                                                 & 102                                    & 989                                       & 1387                                        & 6.8k                                                                                            \\
Chronic Fatigue Syndrome                                                       & 998                                                                                          & 139.50                                                                                                 & 162                                    & 1034                                      & 1292                                        & 31.1k                                                                                           \\
IBS                                                                            & 998                                                                                          & 118.70                                                                                                 & 71                                     & 987                                       & 1337                                        & 77.1k                                                                                           \\
Narcolepsy                                                                     & 997                                                                                          & 148.65                                                                                                 & 121                                    & 1311                                      & 1547                                        & 18.9k                                                                                           \\
Bulimia                                                                        & 996                                                                                          & 122.99                                                                                                 & 46                                     & 761                                       & 1316                                        & 32.8k                                                                                           \\
Hypothyroidism                                                                 & 995                                                                                          & 125.91                                                                                                 & 111                                    & 1585                                      & 2088                                        & 35.2k                                                                                           \\
Costochondritis                                                                & 995                                                                                          & 116.97                                                                                                 & 98                                     & 1136                                      & 1488                                        & 8.8k                                                                                            \\
Hyperhidrosis                                                                  & 994                                                                                          & 97.21                                                                                                  & 184                                    & 1076                                      & 1245                                        & 25k                                                                                             \\
Sinusitis                                                                      & 991                                                                                          & 135.45                                                                                                 & 136                                    & 1242                                      & 1979                                        & 5.9k                                                                                            \\
Psychosis                                                                      & 984                                                                                          & 122.91                                                                                                 & 53                                     & 932                                       & 933                                         & 39.8k                                                                                           \\
Thyroid Cancer                                                                 & 976                                                                                          & 121.80                                                                                                 & 143                                    & 1157                                      & 1405                                        & 3.2k                                                                                            \\
Cystic Fibrosis                                                                & 970                                                                                          & 96.11                                                                                                  & 77                                     & 1001                                      & 882                                         & 7.1k                                                                                            \\
POTS                                                                           & 963                                                                                          & 111.03                                                                                                 & 77                                     & 1155                                      & 1274                                        & 21.8k                                                                                           \\
Multiple Sclerosis                                                             & 958                                                                                          & 152.47                                                                                                 & 129                                    & 1081                                      & 1309                                        & 31.6k                                                                                           \\
Gout                                                                           & 933                                                                                          & 128.87                                                                                                 & 154                                    & 1251                                      & 1730                                        & 14.2k                                                                                           \\
ADHD                                                                           & 899                                                                                          & 222.41                                                                                                 & 141                                    & 875                                       & 1222                                        & 1.4M                                                                                            \\
Gastroparesis                                                                  & 861                                                                                          & 134.91                                                                                                 & 52                                     & 909                                       & 1319                                        & 8k                                                                                              \\
Diabetes (Type I \& II)                                                        & 748                                                                                          & 113.85                                                                                                 & 40                                     & 667                                       & 620                                         & 90.4k                                                                                           \\
Crohn's Disease                                                                & 791                                                                                          & 99.79                                                                                                  & 92                                     & 1026                                      & 995                                         & 43.7k                                                                                           \\
Lupus                                                                          & 784                                                                                          & 93.13                                                                                                  & 96                                     & 978                                       & 972                                         & 18.2k                                                                                           \\
Rheumatoid Arthritis                                                           & 759                                                                                          & 103.08                                                                                                 & 105                                    & 1033                                      & 1010                                        & 6.4k                                                                                            \\
Epilepsy                                                                       & 670                                                                                          & 165.77                                                                                                 & 37                                     & 634                                       & 1170                                        & 27.8k                                                                                           \\
GERD                                                                           & 650                                                                                          & 164.12                                                                                                 & 45                                     & 669                                       & 1518                                        & 44.2k                                                                                           \\
Ankylosing Spondylitis                                                         & 644                                                                                          & 170.83                                                                                                 & 32                                     & 649                                       & 1139                                        & 12.6k                                                                                           \\ \bottomrule
\end{tabular}
}
\caption{Population-wise descriptive statistics.}
\label{tab_apx:stats}
\end{table*}

\begin{table*}[]
\centering
\resizebox{16.0cm}{!}{
\begin{tabular}{@{}cccccccc@{}}
\toprule
\multicolumn{1}{l}{}                                                                                                                                                                                                                       & \multicolumn{1}{l}{}  & \multicolumn{3}{c}{\textbf{Average \# \textit{per population}}}                                                      & \multicolumn{3}{c}{\textbf{Average \# \textit{per claim}}    }                          \\ \cmidrule(l){3-8} 
\textbf{Population type}                                                                                                                                                                                                                            & \textbf{\# Posts}              & \textbf{Questions}                & \textbf{Experiences}              & \textbf{Claims}                                       & \textbf{Populations}           & \textbf{Interventions}         & \textbf{Outcomes}              \\ \midrule
\textbf{Very Common}                                                                                                                                                                                                                                & \multirow{2}{*}{5467} & \multirow{2}{*}{1101.82} & \multirow{2}{*}{1654.00} & \multicolumn{1}{c|}{\multirow{2}{*}{114.83}} & \multirow{2}{*}{0.82} & \multirow{2}{*}{2.66} & \multirow{2}{*}{3.57} \\
\begin{tabular}[c]{@{}c@{}}\small{(Dysthymia, Hypothyroidism, Gout, etc)}\end{tabular}                                                                                                                        &                       &                          &                          & \multicolumn{1}{c|}{}                        &                       &                       &                       \\
\textbf{Common}                                                                                                                                                                                                                                     & \multirow{2}{*}{9539} & \multirow{2}{*}{847.01}  & \multirow{2}{*}{1141.72} & \multicolumn{1}{c|}{\multirow{2}{*}{74.27}}  & \multirow{2}{*}{1.05} & \multirow{2}{*}{2.95} & \multirow{2}{*}{3.22} \\
\begin{tabular}[c]{@{}c@{}}\small{(Chronic Fatigue Syndrome, Bulimia, Psychosis, etc)}\end{tabular} &                       &                          &                          & \multicolumn{1}{c|}{}                        &                       &                       &                       \\
\textbf{Rare}                                                                                                                                                                                                                                       & \multirow{2}{*}{7295} & \multirow{2}{*}{1028.50} & \multirow{2}{*}{1166.25} & \multicolumn{1}{c|}{\multirow{2}{*}{104.75}} & \multirow{2}{*}{0.97} & \multirow{2}{*}{2.79} & \multirow{2}{*}{3.81} \\
\begin{tabular}[c]{@{}c@{}}\small{(Narcolepsy, Hyperhidrosis, Thyroid Cancer, etc)}\end{tabular}                                                                                   &                       &                          &                          & \multicolumn{1}{c|}{}                        &                       &                       &                       \\ \bottomrule
\end{tabular}
}
\caption{For descriptive purposes we categorize conditions into: Very Common ($>$3 million US cases per year), Common ($>$200k US cases per year), and Rare ($<$200k US cases per year). We only include posts that do \textit{not} contain any media (photos/videos). Number of experiences here \textit{include} claims based on personal experiences. Diabetes is included as both Common (Type II) and a Rare (Type I) type.}
\label{tab:stats}
\end{table*}

Table \ref{tab_apx:stats} provides descriptive statistics for all patient populations (that is, subreddits) included in our dataset. 
Dysthymia has the highest number of posts included in our corpus while Ankylosing Spondylitis has the lowest (due to data filtering described above). 
There is substantial variation in the length of the posts written under different subreddits (e.g., in \texttt{r/ADHD} the average post is $\sim$222 tokens, while in \texttt{r/Lupus} it's only $\sim$93 tokens long). 
Similarly, there are variations in the number of questions, claims, and experiences across populations. 
We used subscriber count as a proxy for gauging how active a community on Reddit is.
For instance, \texttt{r/ADHD} has 1.4M subscribers and so can be considered substantially more active than, say, \texttt{r/Psychosis}, which has 39.8K subscribers.  

\begin{table}[]
\small
\centering
\begin{tabular}{lccc}
\cline{2-4}
            & \textbf{P} & \textbf{R} & \textbf{F1}   \\ \hline
Questions   & 0.73      & 0.68   & 0.70 \\
Claims      & 0.47      & 0.40   & 0.43 \\
Experiences & 0.33      & 0.29   & 0.31 \\ \hline
POP         & 0.85      & 0.81   & 0.83 \\
INT         & 0.56      & 0.50   & 0.53 \\
OUT         & 0.48      & 0.37   & 0.42 \\ \hline
\end{tabular}
\caption{Individual annotator labels evaluated against their own ``aggregated'' labels. }
\label{tab:qval3}
\end{table}

\begin{table*}[]
\centering
\resizebox{16cm}{!}{
\begin{tabular}{@{}lccccccccc@{}}
\toprule
\multicolumn{1}{c}{} & \multicolumn{3}{c}{\textbf{Claims}}                                                                 & \multicolumn{3}{c}{\textbf{Experiences}}                                                   & \multicolumn{3}{c}{\textbf{Questions}}                                                     \\ \cmidrule(l){2-10} 
\multicolumn{1}{c}{} & F1                        & P                         & R                                  & F1                        & P                         & R                         & F1                        & P                         & R                         \\ \midrule
CRF \cite{crf}                  & 33.87                     & 35.61                     & 32.29                              & 40.08                     & 40.52                     & 39.64                     & 86.89                    & 85.55                     & 88.27    \\
\texttt{BERT} \cite{devlin-etal-2019-bert}                & \textbf{52.63}            & 58.82                     & 47.61                              & 56.68                     & 59.46                     & 54.33                     & 92.39                     & 88.76                     & 96.34                     \\
\texttt{BioRedditBERT}  \cite{basaldella-etal-2020-cometa}      & 45.16                     & \textbf{70.92}            & 33.29                              & \textbf{59.51}            & \textbf{62.49}            & \textbf{58.92}            & \textbf{93.61}            & \textbf{89.29}            & \textbf{98.37}            \\
\texttt{RoBERTa}    \cite{Liu2019RoBERTaAR}          & \multicolumn{1}{l}{47.05} & \multicolumn{1}{l}{61.53} & \multicolumn{1}{l}{\textbf{38.09}} & \multicolumn{1}{l}{56.81} & \multicolumn{1}{l}{57.11} & \multicolumn{1}{l}{56.52} & \multicolumn{1}{l}{93.06} & \multicolumn{1}{l}{89.01} & \multicolumn{1}{l}{98.34} \\ \bottomrule
\end{tabular}
}
\caption{Additional results from the test set for the task of identifying spans of Claims, Experiences, and Questions.}
\label{tab_apx:task1_res}
\end{table*}

\begin{table*}[]
\centering
\begin{tabular}{@{}lcccccc@{}}
\toprule
              & P     & R     & F1                        & F1$_{\text{\textbf{POP}}}$ & F1$_{\text{\textbf{INT}}}$ & F1$_{\text{\textbf{OUT}}}$ \\ \midrule
\texttt{BERT} \cite{devlin-etal-2019-bert}         & 43.88 & 36.13 & \multicolumn{1}{c|}{39.62} & 41.77   & 44.68   & 33.05   \\
\texttt{RoBERTa} \cite{Liu2019RoBERTaAR}      & \textbf{47.45} & \textbf{39.27} & \multicolumn{1}{c|}{\textbf{42.97}} & \textbf{46.09}   & \textbf{45.99}   & \textbf{36.38}   \\
\texttt{BioRedditBERT} \cite{basaldella-etal-2020-cometa} & 44.44 & 36.55 & \multicolumn{1}{c|}{40.12} & 41.92   & 44.31   & 34.61   \\
\texttt{biomedRoBERTa} \cite{domains} & 38.80 & 21.48 & \multicolumn{1}{c|}{27.66} & 30.54   & 28.13   & 24.54   \\
t5-small \cite{JMLR:v21:20-074} & 41.49 & 38.55 & \multicolumn{1}{c|}{39.97} & 39.61   & 45.02   & 32.41   \\ \bottomrule
\end{tabular}
\caption{Additional results from the test set for the token-level PIO labelling task.}
\label{tab_apx:res-pico}
\end{table*}

\begin{table*}[]
\centering
\resizebox{15cm}{!}{
\begin{tabular}{@{}lcccccccccc@{}}
\toprule
                                                                  & \multicolumn{5}{c}{\textbf{MRR @k}}                                                                     & \multicolumn{5}{c}{\textbf{Precision @k}}                                           \\ \cmidrule(l){2-11} 
\multicolumn{1}{c}{\textbf{k}}                                  & 1              & 5              & 10             & 50             & \multicolumn{1}{c|}{100}            & 1              & 5              & 10             & 50             & 100            \\ \midrule
random                                                            & 0.00           & 0.003          & 0.02           & 0.02           & \multicolumn{1}{c|}{0.02}           & 0.00           & 0.02           & 0.00           & 1.10           & 2.80           \\
BM25                                                              & 5.34           & 7.98           & 9.86           & 14.36          & \multicolumn{1}{c|}{16.70}          & 5.34           & 10.40          & 14.45          & 26.20          & 33.14          \\
DPR \cite{karpukhin-etal-2020-dense}                                                              & 8.07           & 10.96          & 11.89          & 12.20          & \multicolumn{1}{c|}{13.77}          & 8.07           & 16.50          & 23.58          & 31.98          & 36.87          \\ \midrule
\multicolumn{10}{c}{\small{\textit{(trained on the \textbf{\textcolor{red}{Red}HOT} pseudo training set)}}}                                                                                                                                                                           \\
\textbf{\textcolor{red}{Red}HOT}-DER (\texttt{BERT}-based)    & 39.14          & 47.99          & 49.3           & 50.28          & \multicolumn{1}{c|}{50.35}          & 39.14          & 62.55          & 72.64          & 83.73          & 91.74          \\
\textbf{\textcolor{red}{Red}HOT}-DER (\texttt{RoBERTa}-based) & \textbf{45.93} & \textbf{54.60} & \textbf{55.90} & \textbf{56.73} & \multicolumn{1}{c|}{\textbf{56.78}} & \textbf{45.93} & \textbf{69.90} & \textbf{78.81} & \textbf{94.73} & \textbf{98.06} \\ \bottomrule
\end{tabular}
}
\caption{Additional results from the retrieval task (tested on the pseudo test set).}
\label{tab_apx:retrieval-results}
\end{table*}

\section{Additional Results and Experimental Details}
\label{sec_apx:add_results}

We provide results from additional BERT \cite{devlin-etal-2019-bert} variants for the first task of identifying claims, questions, and experiences in Table \ref{tab_apx:task1_res}.
Unsurprisingly, pre-trained neural models consistently outperform linear-chain Bag-of-Words CRFs. 
Similarly, Table \ref{tab_apx:res-pico} provides results from BERT variants and t5-small \cite{JMLR:v21:20-074} for the second task of extracting PICO elements conditioned on the post and a given claim. For the t5 model, the target was to produce \texttt{<entity token>} followed by \texttt{<entity label>} in the same order as they appear in the input sentence (sequential linearization scheme). We evaluated the generated entities against the true sets of PICO elements for each output. While it may be possible to come up with a more optimal linearization scheme for sequence labelling, we posit that to be beyond the scope of our work.   

To use dense retrieval models to rank evidence (abstracts) with respect to their relevance to a given claim we need an efficient means to index  vectors for $\sim$800k abstracts of RCTs in the Trialstreamer database.
We did so using FAISS \cite{8733051} on an Intel Xeon E5-2650 V3 CPU @2.3GHz with 512GB memory. Building an index of dense embeddings for hundreds of thousands passages is highly resource intensive and required roughly 9 hours on two NVIDIA GeForce GTX 1080Ti GPUs. 

To train the dense retriever, we used standard split of train, development, and test sets ($80\%$-$10\%$-$10\%$). 
We trained the two encoders for $40$ epochs with a learning rate of $10^{-5}$ using the Adam optimizer, linear scheduling with warm up, and a dropout rate of $0.1$. 
We parallelized training over multiple-GPUs; it took roughly 40 hours to train the retriever. 
Our best-performing retrieval model was initialized with RoBERTa-base (250M parameters). 
In addition to the results provided in section \ref{sec:evaluation}, we provide additional results for the retrieval task (evaluated on pseudo test set) in Table \ref{tab_apx:retrieval-results}. 

\begin{table*}[]
\centering
\resizebox{16.5cm}{!}{
\begin{tabular}{@{}lllc@{}}
\toprule
               & \multicolumn{1}{c}{\textbf{\begin{tabular}[c]{@{}c@{}}Original w/ PIO placeholders\\ (Template)\end{tabular}}}                                                                                                                                                                                                                 & \multicolumn{1}{c}{\textbf{\begin{tabular}[c]{@{}c@{}}w/ Substituted PIO elements\\ (Pseudo)\end{tabular}}}                                                                                                                                                                                                                                                                                                                                                                                                                                                                                                                                                                                                                                                                                                                                                                                                                                                                                                                              & \textbf{Population}                     \\ \midrule
\textbf{Claim} & Global spread of \textcolor{teal}{{[}OUT{]}} blamed on \textcolor{blue}{{[}INT{]}}                                                                                                                                                                                                                                                                                 & \begin{tabular}[c]{@{}l@{}}Global spread of \textcolor{teal}{Gradual deterioration of renal function} blamed\\ on \textcolor{blue}{cyclophosphamide}\end{tabular}                                                                                                                                                                                                                                                                                                                                                                                                                                                                                                                                                                                                                                                                                                                                                                                                                                                                                                            & \multirow{2}{*}{Lupus}                  \\
\textbf{Post}  & Global spread of \textcolor{teal}{{[}OUT{]}} blamed on \textcolor{blue}{{[}INT{]}}                                                                                                                                                                                                                                                                              & \begin{tabular}[c]{@{}l@{}}\\ Global spread of \textcolor{teal}{Gradual deterioration of renal function} blamed \\ on \textcolor{blue}{cyclophosphamide}\end{tabular}                                                                                                                                                                                                                                                                                                                                                                                                                                                                                                                                                                                                                                                                                                                                                                                                                                                                                                           &                                         \\ \midrule
\textbf{Claim} & \begin{tabular}[c]{@{}l@{}}I'll be starting \textcolor{blue}{{[}INT{]}} soon and have heard/been told it can cause \\ some serious side effects when first starting to take it.\end{tabular}                                                                                                                                                     & \begin{tabular}[c]{@{}l@{}}I'll be starting \textcolor{blue}{solriamfetol} treatment soon and have heard/been told it can \\ cause some serious side effects when first starting to take it.\end{tabular}                                                                                                                                                                                                                                                                                                                                                                                                                                                                                                                                                                                                                                                                                                                                                                                                                                                  & \multirow{2}{*}{Narcolepsy}             \\
\textbf{Post}  & \begin{tabular}[c]{@{}l@{}} \\ I'll be starting \textcolor{blue}{{[}INT{]}} soon and have heard/been told it can cause \\ some serious  side effects when first starting to take it. Because \\ of this, I let my employer know I may have to be out for a day \\ or two during busiest time of the year, and I'm worried I overshared.\end{tabular} & \begin{tabular}[c]{@{}l@{}} \\ I'll be starting \textcolor{blue}{solriamfetol} treatment soon and have heard/been told it can\\ cause some serious side effects when first starting to take it. Because of this, \\ I let my employer know I may have to be out for a day or two during busiest \\ time of the year, and I'm worried I overshared.\end{tabular}                                                                                                                                                                                                                                                                                                                                                                                                                                                                                                                                                                                                                                                                                                &                                         \\ \midrule
\textbf{Claim} & I read that \textcolor{teal}{{[}OUT{]}} could be due to \textcolor{orange}{{[}POP{]}}.                                                                                                                                                                                                                                                                               & \begin{tabular}[c]{@{}l@{}}I read that \textcolor{teal}{hip and lumbar bone mineral density differences} could be due to \\ \textcolor{orange}{Ankylosing Spondylitis}. \end{tabular}                                                                                                                                                                                                                                                                                                                                                                                                                                                                                                                                                                                                                                                                                                                                                                                                                                                                                           & \multirow{2}{*}{Ankylosing Spondylitis} \\ 
\textbf{Post}  & \begin{tabular}[c]{@{}l@{}}I'm 40M with \textcolor{orange}{{[}POP{]}} and UC and my annual blood work just came \\ back \textcolor{teal}{{[}OUT{]}} (around 2.5). However, my other blood levels are all \\ fine, I eat well, am relatively thin (BMI 24), exercise a lot. I read \\ that \textcolor{teal}{{[}OUT{]}} could be due to \textcolor{orange}{{[}POP{]}}.\end{tabular}                          & \begin{tabular}[c]{@{}l@{}} \\ I'm 40M with \textcolor{orange}{Ankylosing Spondylitis} and UC and my annual blood work just came \\ back \textcolor{teal}{hip and lumbar bone mineral density differences} (around 2.5). However, my \\ other blood levels are all fine, I eat well, am relatively thin (BMI 24), exercise a lot. \\ I read that \textcolor{teal}{hip and lumbar bone mineral density differences} could be due \\ to \textcolor{orange}{Ankylosing Spondylitis}.\end{tabular}                                                                                                                                                                                                                                                                                                                                                                                                                                                                                                                                                                                                                           &                                         \\ \midrule
\textbf{Claim} & \begin{tabular}[c]{@{}l@{}}Surprising I'm seeing research articles that \textcolor{blue}{{[}INT{]}} causes \textcolor{teal}{{[}OUT{]}} \\ past the duration of the drug\end{tabular}                                                                                                                                                                               & \begin{tabular}[c]{@{}l@{}}$\ast$ Surprising I'm seeing research articles that \textcolor{blue}{quetiapine versus aripiprazole} causes \\ \textcolor{teal}{psychopathology, cognition, health-related quality of life, and adverse events} past \\ the duration of the drug.\\ \\ $\diamond$ Surprising I'm seeing research articles that \textcolor{blue}{IPS} causes \textcolor{teal}{levels of stress} past the \\ duration of the drug\\ \\ $\bullet$ Surprising I'm seeing research articles that \textcolor{blue}{olanzapine} causes \textcolor{teal}{discontinuation} \\ \textcolor{teal}{rate} past the duration of the drug\end{tabular}                                                                                                                                                                                                                                                                                                                                                                                                                                                                                                                                            & Psychosis                               \\
\textbf{}      & \multicolumn{3}{l}{}                                                                                                                                                                                                                                                                                                                                                                                                                                                                                                                                                                                                                                                                                                                                                                                                                                                                                                                                                                                                                                                                                                                                                                                                                                                                                                                                                                                                \\
\textbf{Post}  & \begin{tabular}[c]{@{}l@{}}Surprising I'm seeing research articles that \textcolor{blue}{{[}INT{]}} causes \textcolor{teal}{{[}OUT{]}} \\ past the duration of the drug. I only took a brief look into it. Has \\ anyone here had \textcolor{blue}{{[}INT{]}} induced \textcolor{teal}{{[}OUT{]}}? What is r/psychosis \\ experience with \textcolor{blue}{{[}INT{]}}?\end{tabular}                                      & \begin{tabular}[c]{@{}l@{}} $\ast$ Surprising I'm seeing research articles that \textcolor{blue}{quetiapine versus aripiprazole} causes \\ \textcolor{teal}{psychopathology, cognition, health-related quality of life, and adverse events} past \\ the duration of the drug. I only took a brief look into it. Has anyone here had \textcolor{blue}{quetiapine} \\ \textcolor{blue}{versus aripiprazole} induced \textcolor{teal}{psychopathology, cognition, health-related quality of life, and} \\ \textcolor{teal}{adverse events}? What is r/psychosis experience with \textcolor{blue}{quetiapine versus aripiprazole?}\\ \\ $\diamond$ Surprising I'm seeing research articles that \textcolor{blue}{IPS} causes \textcolor{teal}{levels of stress} past the \\ duration of the drug. I only took a brief look into it. Has anyone here had \textcolor{blue}{IPS} induced \\ \textcolor{teal}{levels of stress}? What is r/psychosis experience with \textcolor{blue}{IPS?}\\ \\ $\bullet$ Surprising I'm seeing research articles that \textcolor{blue}{olanzapine} causes \textcolor{teal}{discontinuation rate} \\ past the duration of the drug. I only took a brief look into it. Has anyone here had \\ \textcolor{blue}{olanzapine} induced \textcolor{teal}{discontinuation rate}? What is r/psychosis experience with \textcolor{blue}{olanzapine}?\end{tabular} & \multicolumn{1}{l}{}                    \\ \bottomrule
\end{tabular}
}
\caption{Examples of template claims used for the creation of pseudo training labels for training a supervised evidence retrieval model.}
\label{tab_apx:pseudo_examples}
\end{table*}

\begin{table*}
\centering
\resizebox{16.5cm}{!}{
\begin{tabular}{@{}lcc@{}}
\toprule
                                                                                                                                                                                                                                                                                                                                                                                                                                                                                                                                                                                                                                                                                                                                                                                                                                                                                                                                                                                                                                                                                                                                                                                                                                            & \textbf{Title of trial paper}                                                                                                                                                                                                                                                                                           & \begin{tabular}[c]{@{}c@{}}\textbf{Link to abstract/trial}\end{tabular} \\ \midrule 
\textbf{Claim:} \textcolor{blue}{Vitamin} D may prevent \textcolor{teal}{autoimmune diease}                                                                                                                                                                                                                                                                                                                                                                                                                                                                                                                                                                                                                                                                                                                                                                                                                                                                                                                                                                                                                                                                                                                                                                                           & \multirow{2}{*}{\begin{tabular}[c]{@{}c@{}}Vitamin D and marine omega 3 fatty acid supplementation \\ and incident autoimmune disease: VITAL randomized controlled trial.\end{tabular}}                                                                                                         & \multirow{2}{*}{\url{https://dx.doi.org/10.1136/bmj-2021-066452}}      \\
\begin{tabular}[c]{@{}l@{}}\\\textbf{Post:} Okay so... the only bloodwork for me that was pretty \\ abnormal was \textcolor{blue}{vitamin D}. My neurologist did bloodwork \\ for it a year ago and it was in the 20s. He said it should \\ be 50+ and that \textit{\textcolor{blue}{Vitamin D} may prevent \textcolor{teal}{autoimmune}} \\ \textit{\textcolor{teal}{diease}}. Are there any long term problems I should be \\ aware about if I can‚ how get it to go up?\end{tabular}                                                                                                                                                                                                                                                                                                                                                                                                                                                                                                                                                                                                                                                                                                                                                                                                                                                          &                                                                                                                                                                                                                                                                                                 &                                                                  \\ \midrule
\begin{tabular}[c]{@{}l@{}}\textbf{Claim:} been researching few weeks now and I recently came across \\ \textcolor{orange}{POTS Syndrome}. I found that it affects your \textcolor{teal}{heart rate} so I \\ decided to test mine while \textcolor{blue}{resting and then standing} to see \\ if maybe thats what it could be.\end{tabular}                                                                                                                                                                                                                                                                                                                                                                                                                                                                                                                                                                                                                                                                                                                                                                                                                                                                                                                                                               & \multirow{2}{*}{\begin{tabular}[c]{@{}c@{}}Cardiovascular exercise as a treatment of postural orthostatic tachycardia \\ syndrome: A pragmatic treatment trial.\end{tabular}}                                                                                                                   & \multirow{2}{*}{\url{https://dx.doi.org/10.1016/j.hrthm.2021.01.017}}  \\
\begin{tabular}[c]{@{}l@{}}\\\textbf{Post:} I just joined this group, so I apologize if this is not allowed. I have been \\ researching what I feel to be abnormal symptoms Ive been dealing with \\ the majority of my life (dizziness, nausea when standing, etc).. Anyways, \\ Ive \textit{been researching few weeks now and I recently came across \textcolor{orange}{POTS}} \\ \textit{\textcolor{orange}{Syndrome}. I found that it affects your \textcolor{teal}{heart rate} so I decided to test mine} \\ \textit{while resting and then standing to see if maybe thats what it could be.} I \\ took my \textcolor{teal}{heart rate} three times while \textcolor{blue}{laying in bed}. at 1:37am, by heart rate \\ was 73bpm. at 1:39am, my heart rate was 74bpm. at 1:40am, my heart \\ rate was 73 bpm again. I then \textcolor{blue}{stood up} (right next to my bed) and \\ proceeded to take my heart rate again. Immediately \textcolor{teal}{it shot up to more} \\ \textcolor{teal}{than double my resting heart rate} at 1:41am my heart rate was 156bpm \\ i took it again a minute later and at 1:42am my heart rate was 153bpm. \\ Even if its not pots, just from standing up, I feel like this is not a normal \\ bodily response for the majority of the population. Dont know how to go \\ about getting this checked out. By the way, not sure if it matters, but \\ I am a \textcolor{orange}{19 year old girl}.\end{tabular} &                                                                                                                                                                                                                                                                                                 &                                                                  \\ \midrule
\begin{tabular}[c]{@{}l@{}} \textbf{Claim:} did some research and apparently \textcolor{blue}{smoking} can \textcolor{teal}{effect bowel} \\ \textcolor{teal}{movements (bloating,cramping)} which is what i struggle with exactly\end{tabular}                                                                                                                                                                                                                                                                                                                                                                                                                                                                                                                                                                                                                                                                                                                                                                                                                                                                                                                                                                                                                                                          & \multirow{2}{*}{\begin{tabular}[c]{@{}c@{}}The effect of alpha-tocopherol and beta-carotene supplementation on colorectal \\ adenomas in middle-aged male smokers.\end{tabular}}                                                                                                                & \multirow{2}{*}{\url{https://www.ncbi.nlm.nih.gov/pubmed/10385137}}    \\
\begin{tabular}[c]{@{}l@{}}\\\textbf{Post:} i have been a \textcolor{orange}{smoker} for only 3 years, and i recently had the realisation \\ that my \textcolor{orange}{IBS}(like) symptoms correlated to the same period of time i started\\ \textcolor{blue}{smoking}, i then did some research and apparently \textcolor{blue}{smoking} can \textcolor{teal}{effect} \\ \textcolor{teal}{bowel movements (bloating,cramping)} which is what i struggle with \\ exactly, so i dont know if anyone has a similar story or if \textcolor{blue}{quitting} \\ \textcolor{blue}{smoking} helped with their \textcolor{orange}{IBS} ?\end{tabular}                                                                                                                                                                                                                                                                                                                                                                                                                                                                                                                                                                                                                                                                                                                                                                                 &                                                                                                                                                                                                                                                                                                 &                                                                  \\ \midrule
\textbf{Claim:} I read of the issues it can cause the body but so much out there has it.                                                                                                                                                                                                                                                                                                                                                                                                                                                                                                                                                                                                                                                                                                                                                                                                                                                                                                                                                                                                                                                                                                                                                          & \multirow{2}{*}{\begin{tabular}[c]{@{}c@{}}Glycemic Effects of Rebaudioside A and Erythritol in People with Glucose \\ Intolerance.\end{tabular}}                                                                                                                                               & \multirow{2}{*}{\url{https://dx.doi.org/10.4093/dmj.2016.40.4.283}}    \\
\begin{tabular}[c]{@{}l@{}}\\\textbf{Post:} \textcolor{blue}{sugar alcohol} vs \textcolor{blue}{sugar} Just wondering what your thoughts are of \textcolor{blue}{sugar alcohol}.   \\ I noticed a lot of sugar free foods have sugar alcohol inplace of sugar. I read \\ of the \textcolor{teal}{issues it can cause} the body but so much out there has it. Do you avoid \\ sugar alcohol products or do you embrace it as a sugar alternative?\end{tabular}                                                                                                                                                                                                                                                                                                                                                                                                                                                                                                                                                                                                                                                                                                                                                                                                                                                                  &                                                                                                                                                                                                                                                                                                 &                                                                  \\ \midrule
\textbf{Claim:} I cant help thinking it may be related to my \textcolor{blue}{meds}                                                                                                                                                                                                                                                                                                                                                                                                                                                                                                                                                                                                                                                                                                                                                                                                                                                                                                                                                                                                                                                                                                                                                                                 & \multirow{2}{*}{\begin{tabular}[c]{@{}c@{}}Clinical Observation of Levothyroxine Sodium Combined with Selenium in the \\ Treatment of Patients with Chronic Lymphocytic Thyroiditis \\ and Hypothyroidism and the Effects on Thyroid Function, Mood, \\ and Inflammatory Factors.\end{tabular}} & \multirow{2}{*}{\url{https://dx.doi.org/10.1155/2021/5471281}}         \\
\begin{tabular}[c]{@{}l@{}}\\\textbf{Post:} I stopped taking \textcolor{blue}{Levothyroxin} for about a month. Ever since I started \\ taking it again I feel like \textcolor{teal}{crying} after taking it in the mornings. It could\\  be that I really dont want to go to work, but I cant help thinking it may \\ be related to my \textcolor{blue}{meds}. Does this happen to anyone else?\end{tabular}                                                                                                                                                                                                                                                                                                                                                                                                                                                                                                                                                                                                                                                                                                                                                                                                                                                                                                 &                                                                                                                                                                                                                                                                                                 &                                                                  \\ \bottomrule
\end{tabular}
}
\caption{Examples of evidence abstracts (marked relevant by domain experts) retrieved by the RoBERTa-based \textbf{\textcolor{red}{Red}HOT-DER} model trained on pseduo data.}
\label{tab_apx:retrieved_examples}
\end{table*}

\section{Deriving Pseudo Training Data: Examples}
\label{sec_apx:examples}

Generating pseudo training data --- i.e., matching reddit annotated reddit posts to ``relevant'' abstracts of RCTs --- is an important component of our dense retrieval pipeline. 
In Table \ref{tab_apx:pseudo_examples} we provide several examples of the pseudo data we generated from annotated claims. 
For each row we have inserted intervention and outcome elements from abstracts indexed in Trialstreamer, which makes them ``relevant'' by construction (while still featuring natural language as it used on social media).
We showcase how stage-2 annotated (post, claim) pairs serve as templates to create pseudo claims by substituting PICO elements from an existing corpus.

In Section \ref{sec:human_eval} we emphasize the need to evaluate retrieved evidence relevant to \textit{naturally occurring} medical claims, as opposed to the \textit{pseudo} data we derived for training. To this end, we hired domain experts (medical doctors) to look at the evidence abstracts from our retrieval model and assign a relevance score to each abstract (3: relevant, 2: somewhat relevant, 1: irrelevant). We provide some examples of retrieved evidence in Table \ref{tab_apx:retrieved_examples} annotated by our experts as \textit{relevant} (score: 3). Due to space constraints, we provide a link to the full article instead of the full abstract text.

\end{document}